\documentclass[review]{elsarticle}

\usepackage{lineno,hyperref}
\modulolinenumbers[5]

\journal{Journal of \LaTeX\ Templates}

\usepackage{amsmath,amssymb,amsfonts}
\usepackage{graphicx}
\usepackage{textcomp}
\usepackage{xcolor}
\usepackage{diagbox}
\usepackage{subfigure}
\usepackage{booktabs}
\usepackage{multirow}
\usepackage{makecell}
\usepackage{color}
\usepackage{float}
\usepackage{url}
\usepackage{booktabs}
\usepackage{cleveref}
\usepackage{mathrsfs}

\usepackage{bm}
\usepackage{textcomp} 

\usepackage[linesnumbered,ruled,boxed]{algorithm2e}
\usepackage{algpseudocode}



\begin{document}

\begin{frontmatter}

\title{FederatedNILM: A Distributed and Privacy-preserving Framework for Non-intrusive Load Monitoring based on Federated Deep Learning}


\author{Shuang Dai}
\address{Department of Mathematical Sciences, University of Essex, Colchester, UK}

\author{Fanlin Meng\corref{mycorrespondingauthor}}
\address{Department of Mathematical Sciences, University of Essex, Colchester, UK}
\cortext[mycorrespondingauthor]{Corresponding author}
\ead{fanlin.meng@essex.ac.uk}

\author{Qian Wang}
\address{Department of Computer Science, Durham University, Durham, UK}

\author{Xizhong Chen}
\address{School of Engineering, University College Cork, Cork, Ireland}




\begin{abstract}
Non-intrusive load monitoring (NILM), which usually utilizes machine learning methods and is effective in disaggregating smart meter readings from the household-level into appliance-level consumptions, can help to analyze electricity consumption behaviours of users and enable practical smart energy and smart grid applications. However, smart meters are privately owned and distributed, which make real-world applications of NILM challenging. To this end, this paper develops a distributed and privacy-preserving federated deep learning framework for NILM (FederatedNILM), which combines federated learning with a state-of-the-art deep learning architecture to conduct NILM for the classification of typical states of household appliances. Through extensive comparative experiments, the effectiveness of the proposed FederatedNILM framework is demonstrated.

\end{abstract}


\end{frontmatter}

\linenumbers

\section{Introduction}
With the development of smart grids, smart meters have been increasingly installed, which provides enormous opportunities for various smart energy applications such as smart homes and demand response \cite{de2012state}. Smart meters can record high-resolution energy consumption data and therefore have great potential to provide useful insights into the energy use patterns from the individual household level to the grid level. Non-intrusive load monitoring (NILM) \cite{IFed10}, which usually utilizes machine learning in disaggregating smart meter readings from the household-level into appliance-level, is particularly useful for analyzing detailed energy consumption behaviours and enabling practical smart energy applications especially at the low-voltage level such as energy demand management and local energy trading. 

Existing studies on NILM can be broadly categorized into supervised learning and unsupervised learning approaches. For the former, deep learning based models, which provide new opportunities for the electrical utility industry \cite{MISHRA2020104000}, are the most representative structures applied to NILM. The commonly adopted deep learning models include convolutional neural network (CNN) \cite{kelly} and recurrent neural network (RNN) \cite{yuan39}. For unsupervised learning, combinatorial optimization (CO) \cite{kelly}, factorial Hidden Markov Model (FHMM) \cite{yuan26} \cite{yuan28} and clustering analysis \cite{yuan49} are commonly used for NILM.  \cite{kelly} compared unsupervised learning (CO, FHMM) with supervised learning (CNN) for NILM where results showed that the CNN-based model performed better than unsupervised models. Although deep learning models usually perform well on NILM, such models still face several challenges:

\begin{itemize}
	\item The number of the labelled data generated by a single household is limited, and the size of the training set has a great impact on the effectiveness of the deep learning model. Therefore, it is necessary to collect the labelled data from multiple data sources on the premise of ensuring data security.
	
	\item Different users have different lifestyles and thus have different electricity usage patterns (i.e. data distribution), which put forward higher requirements for the generalization ability of the NILM model.
	
	\item Given the increasing public attention to data privacy and security preservation, it is necessary to satisfy the needs of training models with not only high precision but also reasonable communication efficiency under the premise of ensuring individual data privacy. 
\end{itemize}

To overcome the above challenges, federated learning \cite{dianxin1-Communication} was proposed where private data of individual users do not need to be uploaded to a central server for centralized training. Instead, under the coordination of the central cloud server, each participant can carry out the model training locally and only exchange typical parameters of their local model such as updated gradients \cite{jiaoyu5}. Compared with other privacy-preserving technologies that need to encrypt the original data set, federated learning does not need to collect the original data centrally, therefore model training in this framework does not involve data transmission and public sharing, and can thus help to achieve individual data privacy protection. On the other hand, the wide penetration of the Internet of Things (IoT) in various areas \cite{VEERAMANIKANDAN2020103785} has sparked new opportunities for federated learning by providing massive amounts of distributed user-generated data on intelligent IoT devices and applications, which is poised to make substantial contributions in all aspects of our modern life, such as smart healthcare and smart grid system \cite{atzori2010internet}.

Although federated learning, as a newly proposed distributed and privacy-preserving framework, has been studied in various areas, limited attention has been paid to smart grid applications especially for NILM. Since NILM is one of the key technologies to unlock the full potential of local and distributed energy resources, a distributed and privacy-preserving framework is urgently needed to enable its practical applications. To this end, this paper develops a distributed and privacy-preserving federated deep learning framework for NILM (FederatedNILM). The main contributions of this paper are summarised as follows. 

\begin{itemize}
    
    \item We develop the FederatedNILM, a distributed and privacy-preserving framework for NILM based on federated deep learning. To the best of our knowledge, this is the first work to study the distributed and privacy-preserving NILM based on federated deep learning.

	\item  We combine an advanced deep neural network architecture with federated learning, which could benefit the whole FederatedNILM framework by providing more accurate state inference for multiple appliances on the household level.
	
	\item Through comparative experiments, we verify the effectiveness of the proposed FederatedNILM by investigating key practical characteristics including communication costs and model accuracy.
\end{itemize}

The remainder of this paper is organized as follows. In Section II, existing studies on NILM with privacy-preserving methods and federated learning applications are reviewed. Preliminaries of NILM and federated deep learning are given in Section III. Section IV details the proposed FederatedNILM framework, followed by the performance evaluation in Section V. The conclusion and future work are given in Section VI. 

\section{Related work}

The existing studies on privacy-preserving methods for NILM as well as federated learning for various applications such as IoT are reviewed in this section.

\subsection{NILM and its privacy-preserving methods}

Existing research shows that the analysis of real-time data may expose user's privacy information \cite{zhinengdianbiao1}. Many studies designed privacy protection schemes for NILM, which can be mainly divided into identity privacy and data privacy protection. 

Identity privacy protection refers to implement user identity recognition while hiding the real ID of the users by special mechanisms such as blind signature \cite{mianxiang30}. For instance, \cite{imianxinag12} adopted blind signature to protect the identity privacy of smart meter users. In the proposed scheme, the power company has access to real-time electricity data but does not know the identity of owners so the privacies of users are protected. 

Data privacy protection, on the other hand, is usually achieved by adding noises or encryptions to electricity consumption data. \cite{PfSM} suggested that the load signatures of appliances in individual households could be moderated by home electrical power routing. \cite{Information} studied the sensitive data in smart metering from an information-theoretic perspective. In the paper, the smart meter readings were diversified by the alternative energy source and the real consumption data were filtered by the storage units. \cite{achieving} proposed a fog computing approach based on differential privacy against NILM, which adds noises to the behaviour parameter derived from the FHMM rather than the original consumption data. The proposed scheme achieved a satisfying trade-off between data utility and privacy. \cite{performance} compared four variants of differential privacy in blockchain-based smart metering, and the experiment showed that such mechanisms could provide an effective privacy-preserving scheme. 

\subsection{Federated learning applications}

The emergence of federated learning on the one hand ensures the privacy of the distributed data such as those collected by IoT devices, and on the other hand, solves the problem of data isolation. \cite{jiyujuanji6} proposed an autonomous self-learning distributed system, which combined anomaly detection with federated learning to detect damaged IoT devices. \cite{multi} used federated learning for multi-task network anomaly detection, which improved the training efficiency compared with multiple single-task models as well as preserved the privacy of data. Later, \cite{network} combined federated learning with the deep neural network to solve the same problem in \cite{multi}. Particularly, transfer learning was adopted to reconstruct the model to further improve the anomaly detection performance. 

Although federated learning has been studied in different areas, limited attention has been paid to NILM. As aforementioned, NILM is one key technology to unlock the potential of smart local energy systems, and therefore developing a practical, distributed and privacy-aware framework for NILM is urgently needed. In this paper, we take the first attempt to address the problem by developing a distributed and privacy-preserving deep learning framework for NILM based on federated learning. 

\section{Preliminaries}
In this section, we introduce  essential concepts related to the proposed federated deep learning framework.

\subsection{Non-intrusive load monitoring}

Given the aggregated load $L_t$ at time $t$:
\begin{equation}\label{eq1}
L_t=\sum_{i=1}^Il_t^{i}+\gamma_t
\end{equation} 
the goal of non-intrusive load monitoring (NILM) is to recover the status of $I$ target electrical appliances. $l_t^i$ and $\gamma_t$ denote the load consumption for the $i$-th appliance and the residual/unmonitored load respectively at time $t$. NILM can be formulated as either a classification task or a regression task depending on the status variables of individual electrical appliances we aim to recover.

For the regression task, the NILM model aims to find the approximation, denotes as $F$, of the true relationship between the aggregated household-level consumption ($L_t$) and the appliance-level consumption.  
\begin{equation}\label{eq2}
\pmb{L}=[\hat{l}_t^{1},\hat{l}_t^{2},\ldots, \hat{l}_t^{i}, \ldots,\hat{l}_t^{I}] = F(L_t)
\end{equation} where $\pmb{L}$ is the predicted load consumption sequence of $I$ target electrical appliances at time $t$.

For the classification task, thresholds need to be set for the NILM model to determine the states (e.g. ON/OFF) of each target appliance. The threshold method we choose in this paper is the activation-time thresholding, which could avoid the negative effect of the abnormal spikes during the OFF state to better improve the inference accuracy \cite{kelly}. For the sake of simplicity, we assume that there are two typical states (ON/OFF) for the target appliances, and the state $s_t^i$ for $i$-th appliance at time $t$ is related to its threshold $\lambda^i$
\begin{equation}
s_t^i=
\left\{\begin{matrix}
1,& {l_t^i \geq \lambda^i}\\
0,& {l_t^i < \lambda^i}
\end{matrix} \right.
\end{equation}
where 0 represents the OFF state, and 1 denotes the ON state. Therefore the classification task for NILM can be defined as
\begin{equation}\label{eq3}
\pmb{S}=[\hat{s}_t^{1},\hat{s}_t^{2},\ldots, \hat{s}_t^{i}, \ldots,\hat{s}_t^{I}] = F_s(L_t)
\end{equation} 
where $\hat{s}_t^{i}$ is a binary variable indicating the predicted ON/OFF state of $i$-th electrical appliance at time $t$.

\subsection{Federated deep learning}

When data owners intend to combine their local data to train a deep neural network model, the traditional way is to integrate their individual data into the central server and use the integrated data to train a standard centralized model. However, the data uploading and integration process often involves legal issues such as data privacy and security. Therefore federated learning was introduced to overcome the problem \cite{li2019federated}.

\begin{figure}[!htbp]
	\centering
	\includegraphics[width=0.8\textwidth]{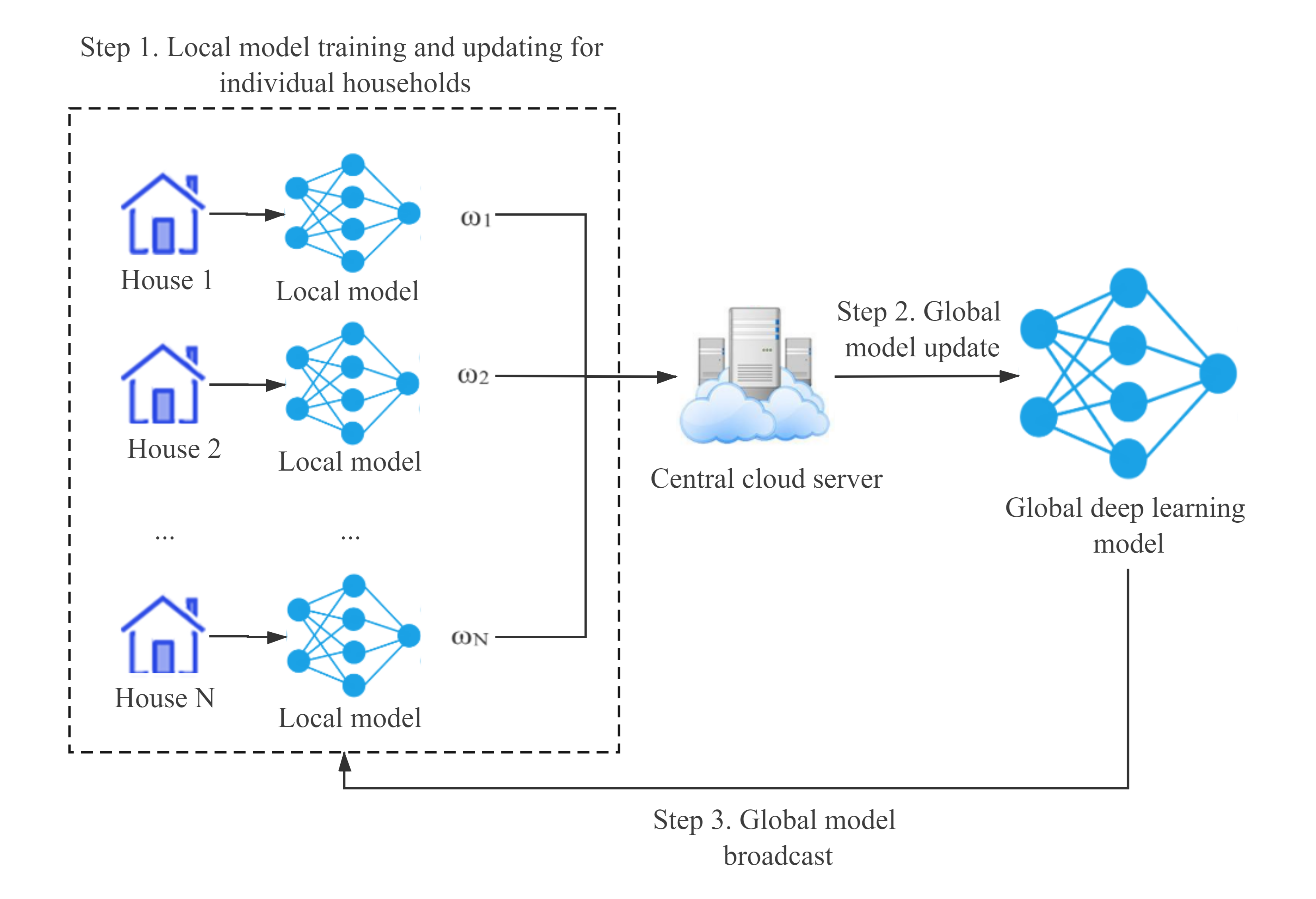}
	\caption{The training process of the federated deep learning framework}
	\label{F1}
\end{figure}

Federated learning is a machine learning strategy whose goal is to train a high-quality global model while the original training data are distributed in each local client without the need of transferring to the central server. The training process of the federated deep learning framework is shown in Figure \ref{F1}. 
As described in the figure, at step 1, each client (i.e. each household) uses its local private data to train the local model and update its local parameters during each training round. Then, each client passes the updated parameters to a central cloud server. At step 2, the global model aggregates the updated parameters from all local clients and updates its own parameters accordingly in the central cloud server. Subsequently, the updated global model parameters are broadcast to each local client at step 3. The process is iterated for multiple rounds until convergence.

Compared to traditional centralized methods, federated learning only requires the transfer of model parameters rather than original data between clients and the central server, which avoids the need of transferring distributed data and therefore is particularly useful for distributed data privacy protection. 

\section{Federated deep learning NILM (FederatedNILM) framework}

In this section, we will first describe the workflow of the proposed FederatedNILM framework, and then detail the architecture of the deep learning network model used in both local clients and the global model.

\begin{algorithm}[!t]\label{A2}  
	\LinesNumbered 
	\scriptsize
	\caption{FederatedNILM for multiple households} 
	\KwIn{Aggregated load consumption of target appliances from all $N$ houses $\left\{L_n\right\}, n=1,2,\ldots,N$, the number of global communication rounds $R_G$, the global sharing batch $B_G$, the local batch $B_L$, the number of local epochs $R_L$}
	
	\KwOut{The optimal global deep learning model parameters $w^*_G$.} 
	\textbf{Initialization:}\\
	The initial global deep learning model parameters $w_G$, learning rate $\eta$, momentum $\rho$, loss function $\mathcal L$, the global sharing batch $B_G$\; 
	The global communication round index $r_G=1$\;
	The local epoch index $r_L=1$\;
	The moment estimate variable $v=0$\;

	\textbf{Procedure:}\\
	// Global deep learning model aggregation, training and broadcasting\\
	\For{$r_G\leq{R_G}$} 
	{
		
		\For{$n=1,2,\ldots,N$ \textbf{in parallel}}{
			$w_{r_G+1}^n \leftarrow HouseholdsUpdate(w_{r_G}^n)$\;
		}
		$w_{r_G+1} \leftarrow \frac{\sum_{n=1}^Nw_{r_{G}+1}^n}{N}$\;
		Replace the old global deep learning model with the new parameters, which are stored in the global sharing batch $B_G$: $w_{r_G} \leftarrow w_{r_G+1}$\;
		$r_G \leftarrow r_G+1$.\
	}
	\textbf{return} The global deep learning model with parameters $w_{R_G}$\\
	// Local households model updating, training and uploading\\
	
	$HouseholdsUpdate(w_{r_G}^n)$:\\
	Split $L_n$ into batches of size $B_L$\; 
	
	\While{${r_L}\le R_L$}{
		\For{each batch of $L_n$}{
			Calculate the gradient by $d\leftarrow\triangledown{w_{r_G}^n}\mathcal L$\;
			Update biased moment estimate variable by $v \leftarrow \rho v + d$\;
			Update the local model parameters by $w_{r_G}^n \leftarrow w_{r_G}^n - \eta v$\; 
		}
	}
	\textbf{return $w_{r_G}^n$}\\
	
\end{algorithm}

\subsection{The workflow of the proposed FederatedNILM framework}

The goal of the FederatedNILM framework is to accurately infer the ON/OFF states of multiple appliances in individual households in a distributed and privacy-preserving way. The whole workflow of the FederatedNILM can be described in three stages (see Algorithm \ref{A2}).

\subsubsection{Initialisation} The parameters of global deep learning model $w_G$, global sharing batch $B_G$, along with parameters of model training including learning rate $\eta$, momentum $\rho$ and loss function $\mathcal L$ need to be initialised first. Then set the global and local communication round index to 1, initialise the moment estimate variable $v$ to 0, and begin the first round of training.

\subsubsection{Local households model updating, training, and uploading} The local households will train their own local deep learning model based on the local data $L_n$ after receiving the broadcast parameters from the global deep learning model. As described in $HouseholdsUpdate()$, in each local epoch, the local deep learning models aim to find the best approximation $F_s$ (Equation (\ref{eq3})). The local training returns the updated parameters from all the local models, which are then uploaded to the global cloud server.

\subsubsection{Global deep learning model aggregation, training, and broadcasting}

The global deep learning model receives the local updated parameters from the global sharing batch $B_G$, and adopts federated averaging (FedAvg) to the parameter sets \cite{dianxin1-Communication}. Then, the global deep learning model will be updated based on the FedAvg results. After this, the updated global parameters will be broadcast to the local models of each house.
After running all the global communication rounds between the local households and the central cloud server, a final global deep learning model will be generated.

\subsection{The deep learning model for NILM} \label{CNNModel}
\begin{figure}[!htbp]
	\centering
	\includegraphics[width=0.73\textwidth]{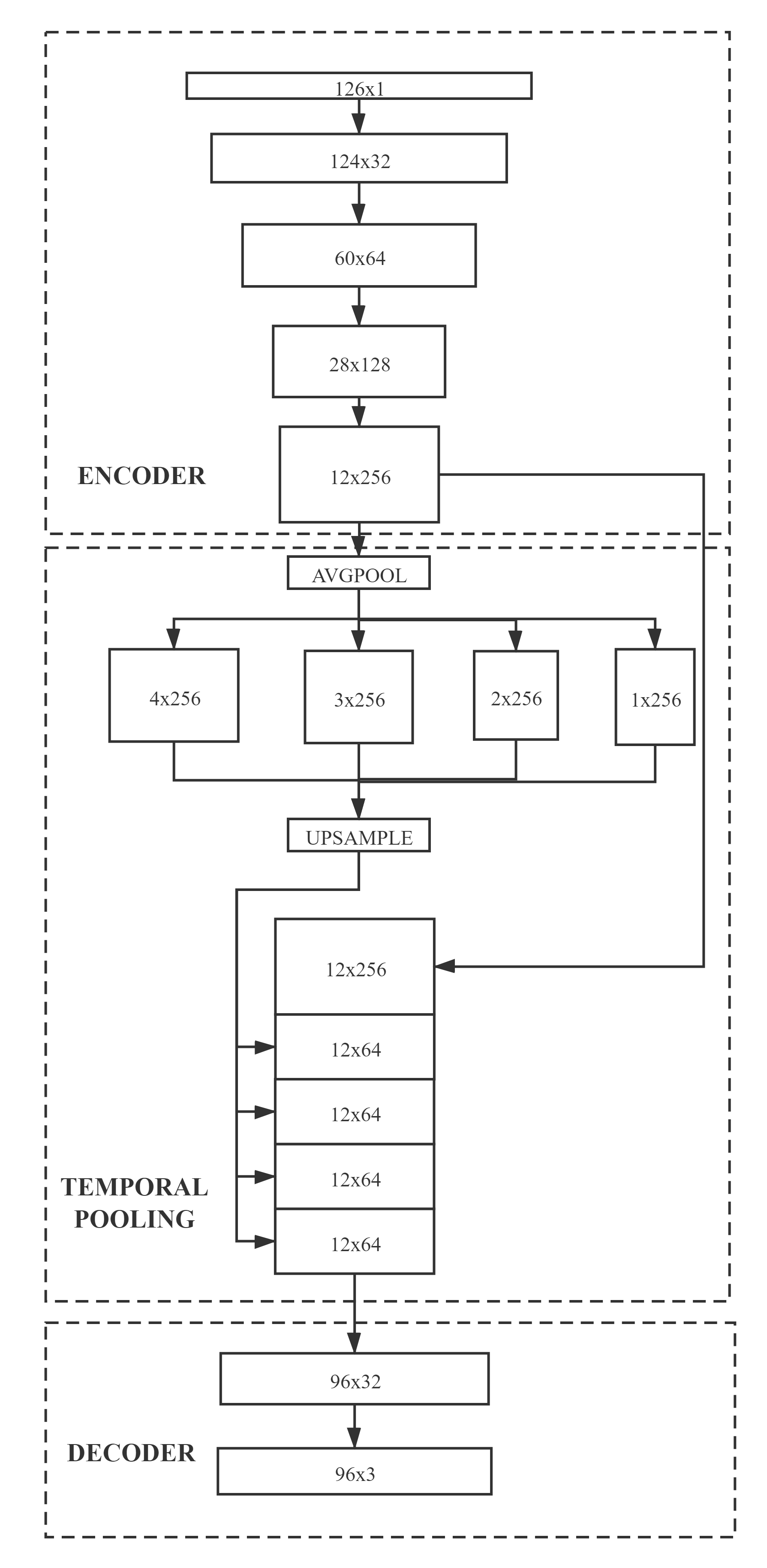}
	\caption{The overall layout of the deep learning model for NILM}
	\label{F2}
\end{figure}

The deep learning architecture utilized to enhance the overall inferring performance is inspired from \cite{PSPNet}, which was originally used for image semantic segmentation. The selection of the above particular architecture is motivated by its potentially promising performance on NILM after appropriate adjustments as demonstrated in \cite{2010.}. The complete layout of our chosen deep learning model for NILM is shown in Figure. \ref{F2}

Specifically, the architecture of the deep learning model for NILM is composed of three modules: the encoder, the temporal pooling module, and the decoder.

\subsubsection{Encoder} The input of the encoder is the household aggregated load consumption of the target appliances over a 12.6 minutes interval (the length of the input window of consumption datasets is 126). The encoder increases the output features from a single aggregation value to 256, while paying the price of decreasing the time signal resolution by 10 times. 

\subsubsection{Temporal pooling} The temporal pooling module consists of four average pooling modules. The filters in this module are reduced from the whole size of the input signal to one-sixth of it, which is the same case with the stride. After going through a convolutional layer, the feature dimension of the input is reduced to a quarter of its original size, and the acquired feature maps were upsampled to increase their size to the size of the input time signals. Then the upsampled feature maps (shallow features) are concatenated with the original input signal (deep features) from the temporal pooling to get the final feature maps. The fusion of the deep and shallow features of the temporal pool could enable this block to get contextual information fed into the decoder.

\subsubsection{Decoder} The decoder receives the output from the temporal pooling block and passes it to a convolutional layer to recover the temporal resolution. Then the output is fed into the final convolutional layer to produce the final output. 

$Softmax$ is utilized to classify the states of multiple targeted appliances. After this, the binary cross-entropy is chosen as the loss function, which is given by:

\begin{equation}\label{eq8}
\begin{split}
\mathcal L_{class}^{i} =& \frac{1}{N}\sum_{n=1}^N\frac{1}{S}\sum_{s=1}^S(s_{n,s}^{i}\cdot\log p(s_{n,s}^{i}) \\&+(1- s_{n,s}^{i})\cdot\log(1-p(s_{n,s}^{i})))
\end{split}
\end{equation} where $N$ is the number of the training samples, $S$ is the length of the output from $Softmax$ classifier, $s_{n,s}^i$ is the true state of $i$-th appliance, and $p(s_{n,s}^{i})$ denotes the predicted probability that $i$-th appliance is in the activation state. Also, stochastic gradient descent (SGD) is selected as the optimizer to facilitate the convergence of the binary cross entropy.

\section{Performance evaluation}

This section gives numerical experiments to evaluate the proposed FederatedNILM framework. Firstly, experimental settings including the environment setup, data preprocessing, training and testing split, and the evaluation metrics are introduced. Secondly, we compare our proposed FederatedNILM with the centralized counterpart in terms of model performance and training time efficiency. Finally, we compare adopted deep learning architecture (used in the FederatedNILM and the centralized counterpart) with several existing advanced NILM models to evaluate the performance of the proposed FederatedNILM framework. 

\subsection{Experimental settings}
\subsubsection{Environmental setup} The FederatedNILM model is implemented on Pytorch and conducted on a Windows 10 platform using an NVIDA GeForce RTX 2080 Ti GPU with 64GB of RAM and CUDA v10.2.

\subsubsection{Dataset and preprocessing} In this paper, we consider the UK-DALE dataset \cite{yuan54}, which is a public and commonly used dataset and recorded both the aggregated load (high frequency and low frequency) and the individual appliances (low frequency) consumption from 5 different households (see Table \ref{t2}).

\begin{table}[!htbp]
	\centering
	\setlength{\tabcolsep}{0.9mm}{
		\setlength{\abovecaptionskip}{0pt}%
		\setlength{\belowcaptionskip}{10pt}%
		\caption{Dataset details}
		\begin{center}
			\begin{tabular}{llllll}
				\toprule
				\hline
				House No.       & 1       & 2       & 3  & 4       & 5       \\\midrule[1pt] \hline
				Total days      & 1631    & 237     & 42 & 208     & 139     \\ \hline
				Appliances      & 52      & 19      & 4  & 11      & 24      \\ \hline
				Fridge          & $\surd$ & $\surd$ & -  & $\surd$ & $\surd$ \\ \hline
				Dishwasher      & $\surd$ & $\surd$ & -  & -       & $\surd$ \\ \hline
				Washing machine & $\surd$ & $\surd$ & -  & $\surd$ & $\surd$ \\ \hline
				
			\end{tabular}
			\label{t2}
	\end{center}}
\end{table}

To demonstrate the effectiveness of the proposed FederatedNILM model and the utilized deep neural network architecture, we deploy a simplified experimental setup following \cite{kelly} for comparison purpose. We consider common appliances possessed by most houses, i.e. fridge, dishwasher, and washing machine, which narrowed the dataset to houses 1, 2 and 5. The date range selected from the three houses is from 12/04/2013 to 01/07/2015 for house 1, 22/05/2013 to 03/10/2013 for house 2, and 29/06/2014 to 01/09/2014 for house 5.

Following the same data preprocessing procedure in \cite{kelly}, abnormal load consumption records were firstly filtered out by the max power threshold provided in Table \ref{t3} followed by a down-sampling of the aggregated load from 1s to 6s to align with the submetered data. The resampled data were normalized by subtracting the mean and then dividing a constant value 2000 W. Finally, the state series of each target appliance were derived from the activation-time thresholding as the input of the proposed FederatedNILM model where relevant thresholds are provided in Table \ref{t3}.

\begin{table}[!htbp]
	\centering
	\setlength{\tabcolsep}{0.9mm}{
		\setlength{\abovecaptionskip}{0pt}%
		\setlength{\belowcaptionskip}{10pt}%
		\caption{Relevant thresholds information}
		\begin{center}
			\begin{tabular}{llll}
				\toprule
				\hline
				& Fridge & Dishwasher & Washing Machine \\ \midrule \hline
				Max power (W)         & 300    & 2500       & 2500            \\ \hline
				Power threshold (W)   & 50     & 20         & 20              \\ \hline
				Min. ON duration (s)  & 1      & 60         & 60              \\ \hline
				Min. OFF duration (s) & 0      & 60         & 5               \\ \hline
				
			\end{tabular}
			\label{t3}
	\end{center}}
\end{table}

\subsubsection{Training and testing}

Both the seen house case and the unseen house case are considered to verify the effectiveness of FederatedNILM. 

For the seen house case, the split of the three houses datasets is listed in Table \ref{tseen}. The first 80\% series from each household were selected as the training set, followed by a 10\% for validation, and 10\% for testing. In this case, the disaggregation ability of the model is evaluated when signatures of specific appliances are learned.

\begin{table}[!htbp]
	\centering
	\setlength{\tabcolsep}{0.9mm}{
		\setlength{\abovecaptionskip}{0pt}%
		\setlength{\belowcaptionskip}{10pt}%
		\caption{Training, validating and testing splits for the seen house case model}
		\begin{center}
			\begin{tabular}{llll}
				\toprule
				\hline
				& Train & Validation & Test \\\midrule \hline
				House 1    & 80\% & 10\%       & 10\% \\ \hline
				House 2      & 80\%  & 10\%        & 10\%  \\ \hline
				House 5      & 80\%  & 10\%        & 10\%  \\ \hline
			\end{tabular}
			\label{tseen}
	\end{center}}
\end{table}

For the unseen house case, the chosen of three houses allows us to train using two houses and test on another. The unseen house case aims to verify the generalization ability of the model, and the generic signature characteristics of the same type of appliances need to be distinguished. As detailed in Table \ref{tunseen}, we split two houses data into training and validation sets, and use the other unseen house as the test set. Then, we average results of all three different combination cases of the training and testing for comparative analysis. 

\begin{table}[!htbp]
	\centering
	\setlength{\tabcolsep}{1mm}{
		\setlength{\abovecaptionskip}{0pt}%
		\setlength{\belowcaptionskip}{10pt}%
		\caption{Training and testing splits for the unseen house case model}
		\begin{center}
			\resizebox{\textwidth}{0.11\textwidth}{ 
            \begin{tabular}{c|ccc|ccc|ccc}
            \toprule
            \hline
            \multirow{2}{*}{\begin{tabular}[c]{@{}c@{}}House \\ Number\end{tabular}} &
              \multicolumn{3}{c|}{Case 1} &
              \multicolumn{3}{c|}{Case 2} &
              \multicolumn{3}{c}{Case 3} \\ \cline{2-10} 
             &
              \multicolumn{1}{l}{Training} &
              \multicolumn{1}{l}{Validation} &
              \multicolumn{1}{l|}{Testing} &
              \multicolumn{1}{l}{Training} &
              \multicolumn{1}{l}{Validation} &
              \multicolumn{1}{l|}{Testing} &
              \multicolumn{1}{l}{Training} &
              \multicolumn{1}{l}{Validation} &
              \multicolumn{1}{l}{Testing} \\ \midrule\hline
            1 & 90\% & 10\% & -     & 90\% & 10\% & -     & -    & -    & 100\% \\ \hline
            2 & 90\% & 10\% & -     & -    & -    & 100\% & 90\% & 10\% & -     \\ \hline
            5 & -    & -    & 100\% & 90\% & 10\% & -     & 90\% & 10\% & -     \\ \hline
            \end{tabular}
			\label{tunseen}}
	\end{center}}
\end{table}

The parameters used in the FederatedNILM model training are listed in Table \ref{t1}. Different global rounds [2, 4, 6, 8, 10] are selected to conduct comparative experiments. We repeat the experiment five times for all the cases and report the average performance for each model.

\begin{table}[!htbp]
	\centering
	\setlength{\tabcolsep}{0.9mm}{
		\setlength{\abovecaptionskip}{0pt}%
		\setlength{\belowcaptionskip}{10pt}%
		\caption{Parameters used in FederatedNILM}
		\begin{center}
			\begin{tabular}{lll}
				\toprule
				\hline
				Item                & Explanation                 & Value                \\ \midrule[1pt] \hline
				$B_G$               & Global sharing batch size   & 32                   \\ \hline
				$R_G$               & Global communication rounds & {[}2, 4, 6, 8, 10{]} \\ \hline
				$B_L$               & Local batch size            & 32                   \\ \hline
				$R_L$               & Local epochs  & 10                   \\ \hline
				$\eta$              & Learning rate               & 1e-4                 \\ \hline
				Activation function & -                           & ReLU                 \\ \hline
				Dropout probability & -                           & 0.1                  \\ \hline
				$\rho$              & Momentum                    & 0.5                  \\ \hline
				Optimizer           & -                           & SGD                  \\ \hline
				
			\end{tabular}
			\label{t1}
	\end{center}}
\end{table}

\subsubsection{Evaluation criteria} Four evaluation metrics are used to assess the performance of the proposed framework.

By denoting true positive as TP, true negative as TN, false positive as FP, and false negative as FN, the evaluation metrics are defined as follows:
\begin{equation}\label{eq9}
Precision = \frac{TP}{TP+FP}
\end{equation}

\begin{equation}\label{eq10}
Recall = \frac{TP}{TP+FN} \\
\end{equation}

\begin{equation}\label{eq11}
Accuracy = \frac{TP+TN}{TP+TN+FP+FN}\\
\end{equation}

\begin{equation}\label{eq12}
F_1 = 2 \times \frac{Precision\times Recall}{Precision + Recall}
\end{equation} 
where precision represents the proportion of TPs to all the data sequences classified to the ON state. Recall denotes the ratio of TPs to all data sequences that are actually in the ON state. Accuracy reflects the ratio of all correctly identified samples to all the data sequences. $F_1$ is defined as a weight average representation for precision and recall within the range of $[0, 1 ]$. An $F_1$ close to 1 indicates that the classification results for the target appliance are better.

\subsection{Comparative studies}
We consider FederatedNILM model and the centralized counterpart (termed as Centralized-NILM) for both the seen house case and unseen house case.

For Centralized-NILM, each household needs to upload their original data directly to the trusted central server, in which the data are trained in a centralized way by the deep learning model described in Section \ref{CNNModel}. This scenario could not provide any data privacy guarantee to participants.

FederatedNILM, on the other hand, does not require access to the original data. In this scenario, households only need to share the training outcomes (parameter sets) of the local deep learning model with the central server. 

\subsubsection{Performance comparison with centralized model based on seen house case} \label{seen}
In this section, we conduct comparative studies between the proposed FederatedNILM and Centralized-NILM for the seen house case. We run 10 global rounds with 10 local epochs for FederatedNILM training, and 100 epochs for Centralized-NILM training.

\begin{figure}[htbp]
\centering

\subfigure[Fridge]{
\begin{minipage}[t]{0.33\linewidth}
\centering
\includegraphics[width=1\textwidth]{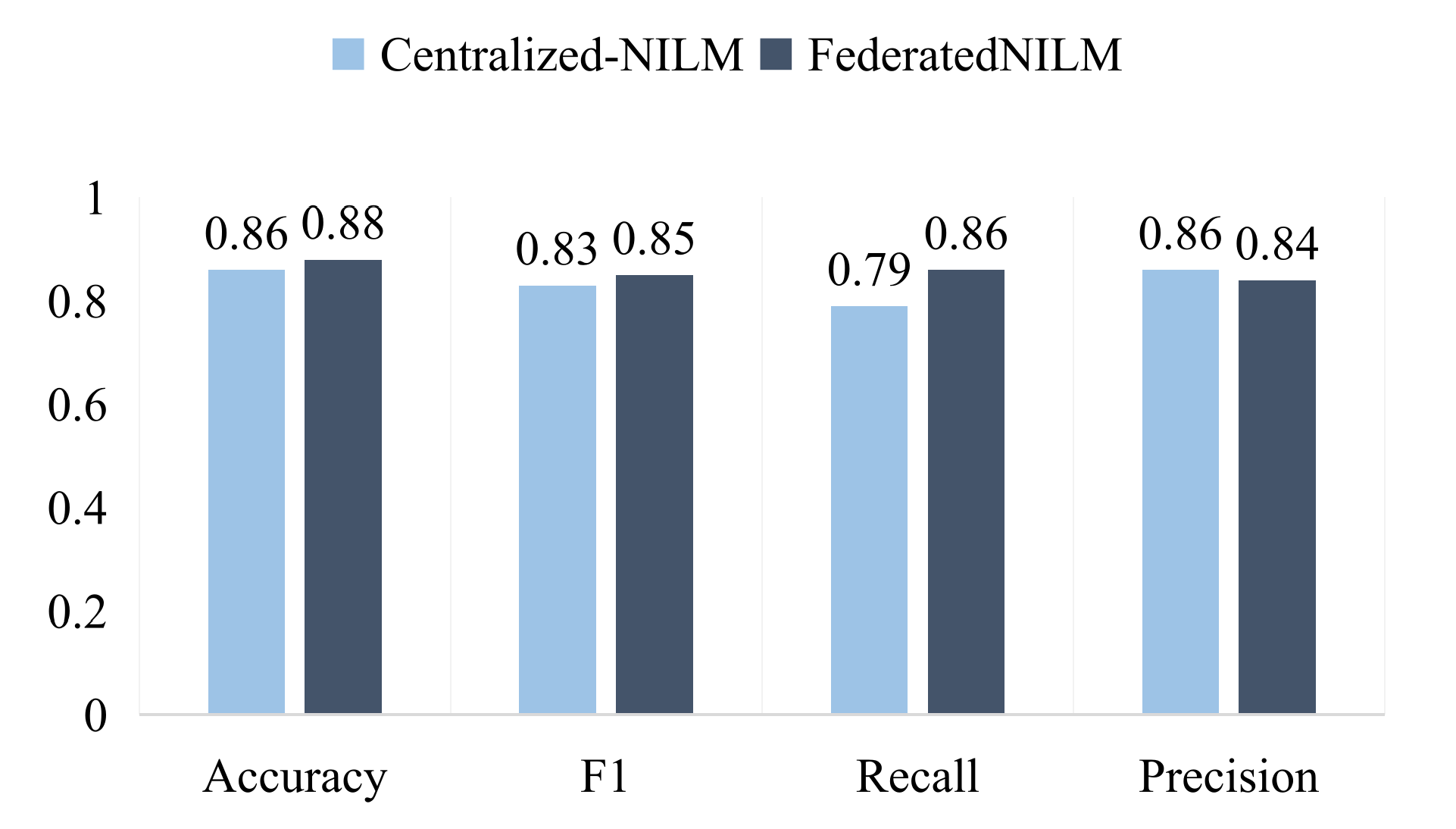}
\end{minipage}%
}%
\subfigure[Dishwasher]{
\begin{minipage}[t]{0.33\linewidth}
\centering
\includegraphics[width=1\textwidth]{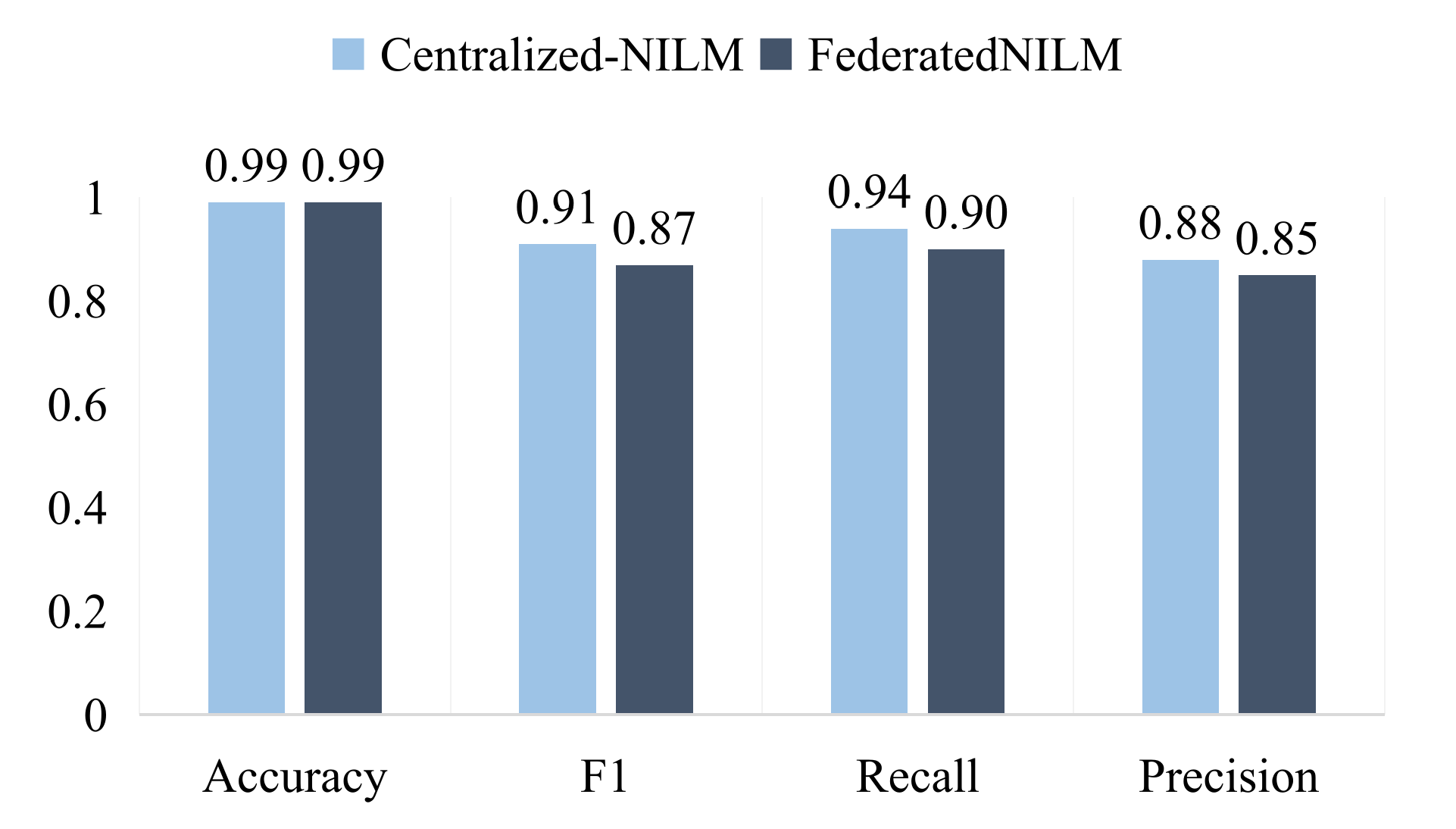}
\end{minipage}%
}%
\subfigure[Washing machine]{
\begin{minipage}[t]{0.33\linewidth}
\centering
\includegraphics[width=1\textwidth]{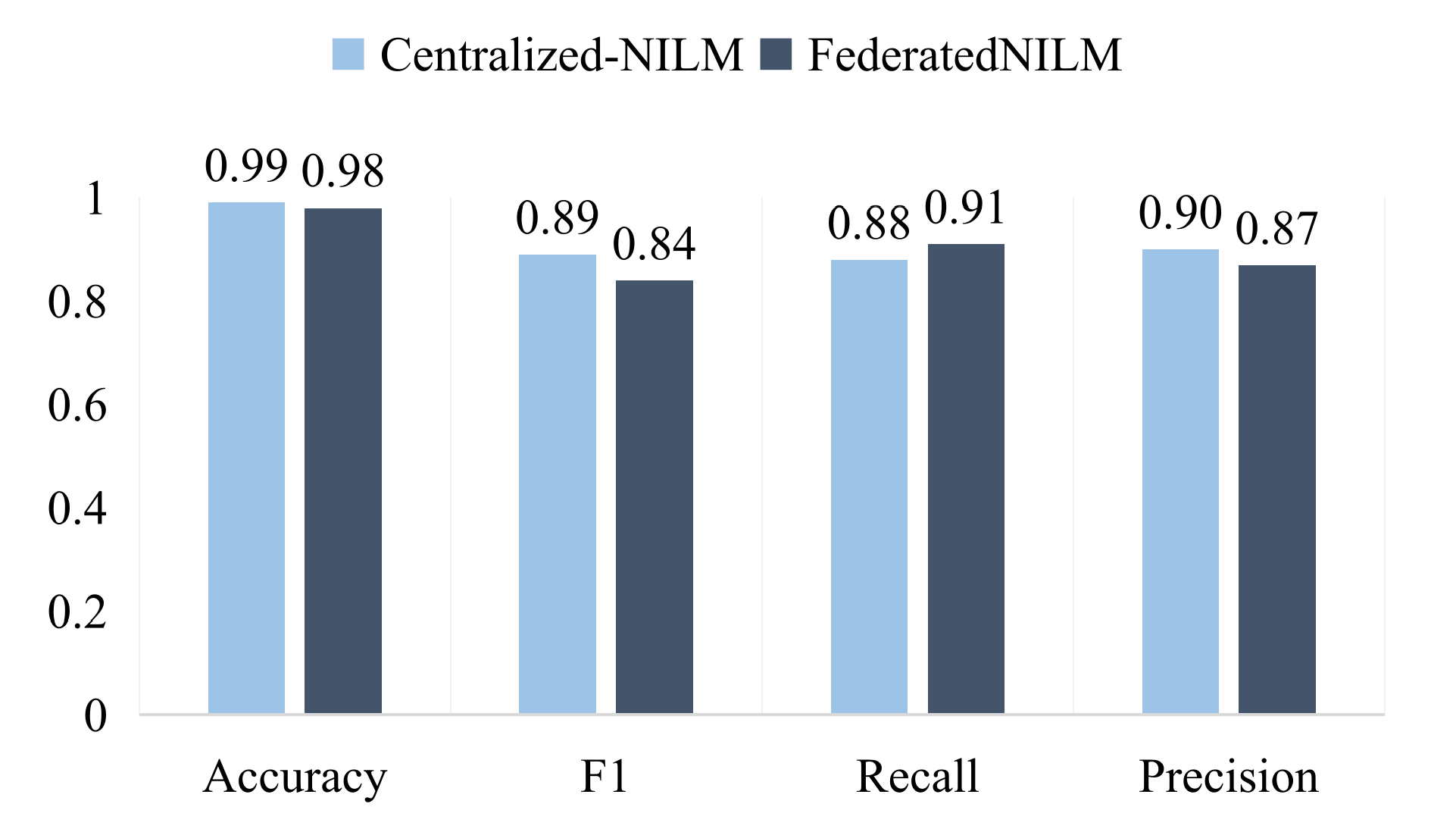}
\end{minipage}
}%
\centering
\caption{ Disaggregation performance in the seen house case.}
\label{Fseen}
\end{figure}

Figure \ref{Fseen} shows the disaggregation performance in the seen house case.

From the results, we can see that for each appliance, both Centralized-NILM and FederatedNILM achieve satisfactory results on the dishwasher and washing machine, and reasonable results on the fridge. It is worth pointing out that the fridge consumes relatively low power compared with other appliances and is likely to be learned with less evident signature during model training since its consumption can easily be omitted as unidentified load noise. This might explain why testing results of the fridge show relatively low scores compared with other appliances. 

It should be highlighted that the FederatedNILM achieves very similar performance to Centralized-NILM on all appliances. Therefore, it is reasonable to infer that our proposed FederatedNILM framework works well in a distributed and privacy-preserving manner for the seen house case and can achieve a good trade-off between data privacy and data utility. 

\subsubsection{Performance comparison with centralized model based on unseen house case} \label{unseen}
In this section, we apply our proposed models for the unseen house case to evaluate the generalization ability of FederatedNILM. 

\begin{figure}[!htbp]
\centering

\subfigure[Fridge]{
\begin{minipage}[t]{0.33\linewidth}
\centering
\includegraphics[width=1\textwidth]{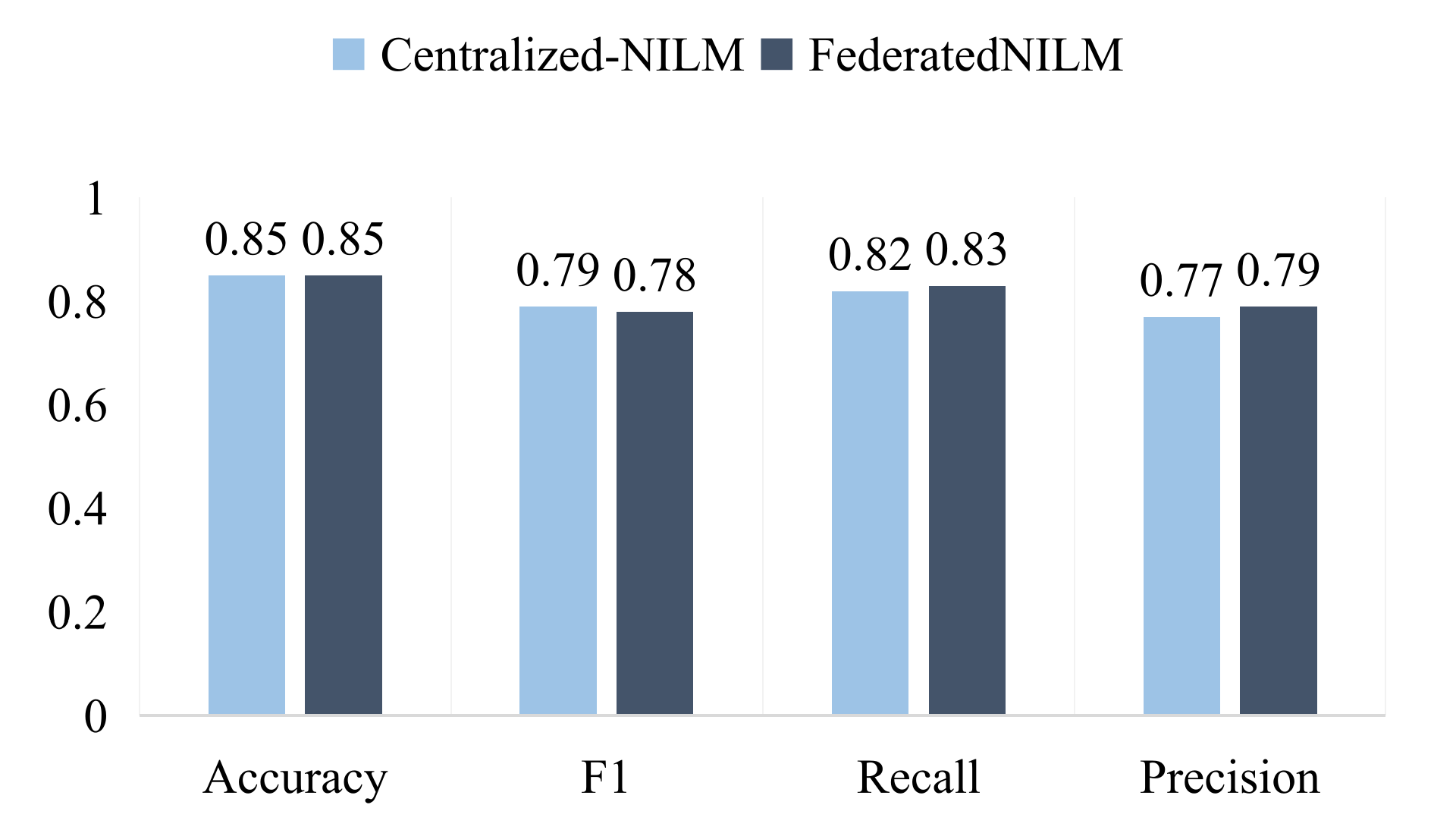}
\end{minipage}%
}%
\subfigure[Dishwasher]{
\begin{minipage}[t]{0.33\linewidth}
\centering
\includegraphics[width=1\textwidth]{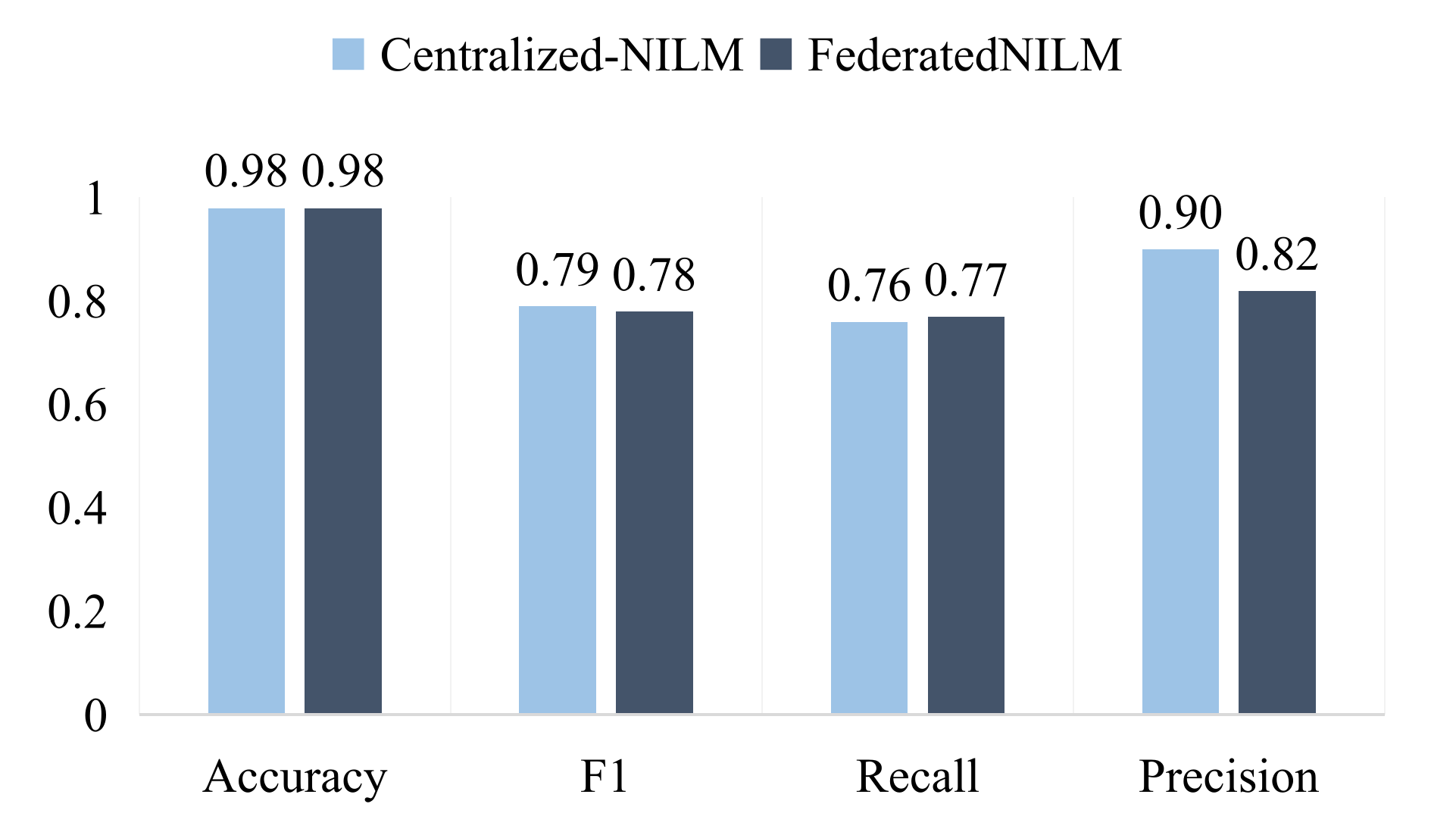}
\end{minipage}%
}%
\subfigure[Washing machine]{
\begin{minipage}[t]{0.33\linewidth}
\centering
\includegraphics[width=1\textwidth]{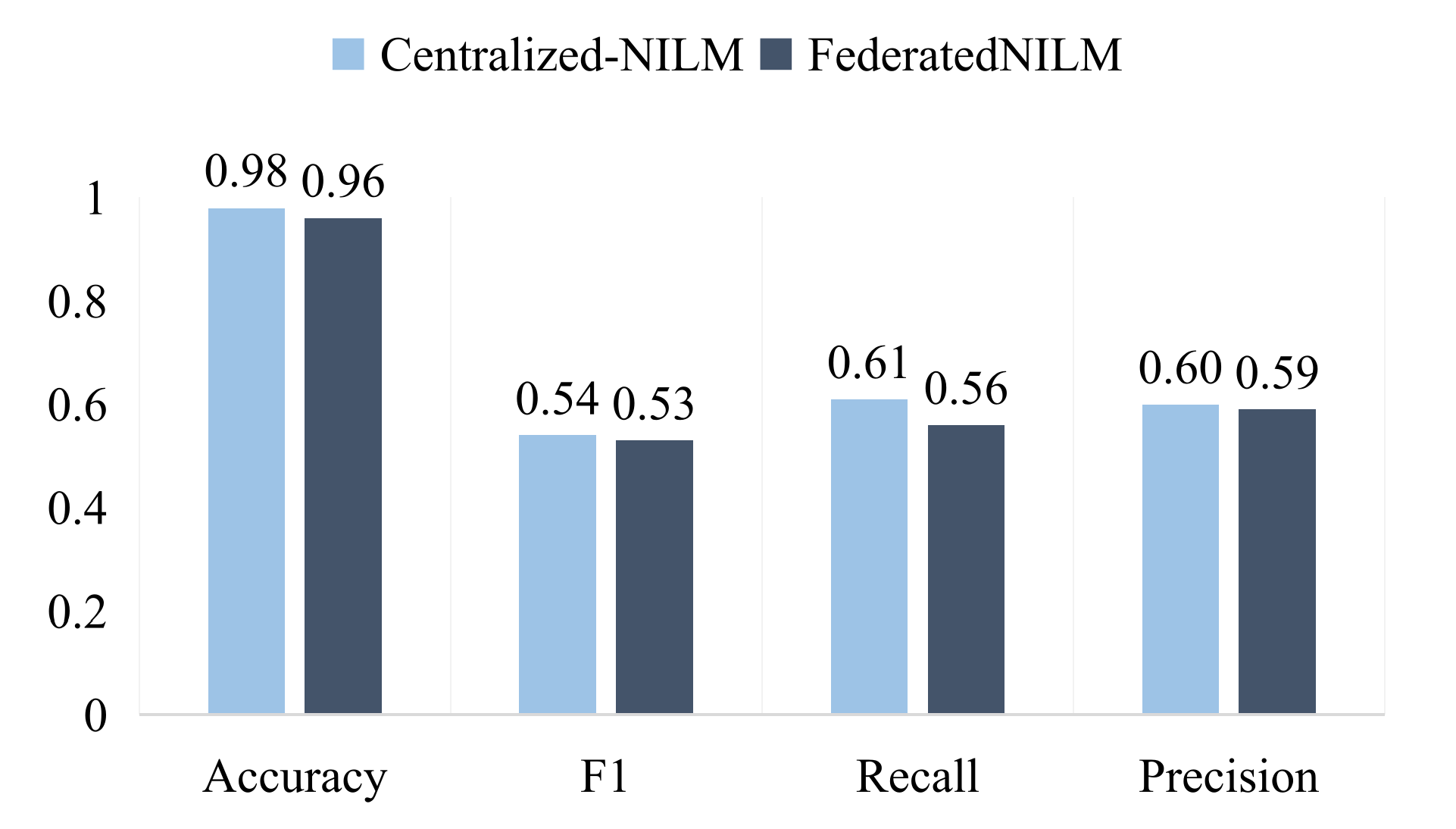}
\end{minipage}
}%
\centering
\caption{ Disaggregation performance in the unseen house case.}
\label{Funseen}
\end{figure}

Figure \ref{Funseen} shows the disaggregation scores on a house not seen during training for the three target appliances.

Compared with the seen house case, both models produce satisfying results for fridge and dishwasher but perform less well for washing machine. One possible reason is that the washing machine, as a multi-state appliance, has more complex signature characteristics \cite{d2019transfer}, which is likely to be more difficult to distinguish the generic signature characteristics. 

We can also find that the FederatedNILM has very similar performance compared with Centralized-NILM on all appliances for the unseen house case, which aligns well with our conclusion on the proposed FederatedNILM framework in terms of data privacy and data utility in Section \ref{seen} for the seen house case. 

\subsubsection{ Global round, local epochs, and training time efficiency} \label{GR}

In this section, different global rounds and their corresponding training costs and testing results are given to explore the relationship between training efficiency and model performance.

\begin{figure}[!t]
	\centering
	\includegraphics[width=1\textwidth]{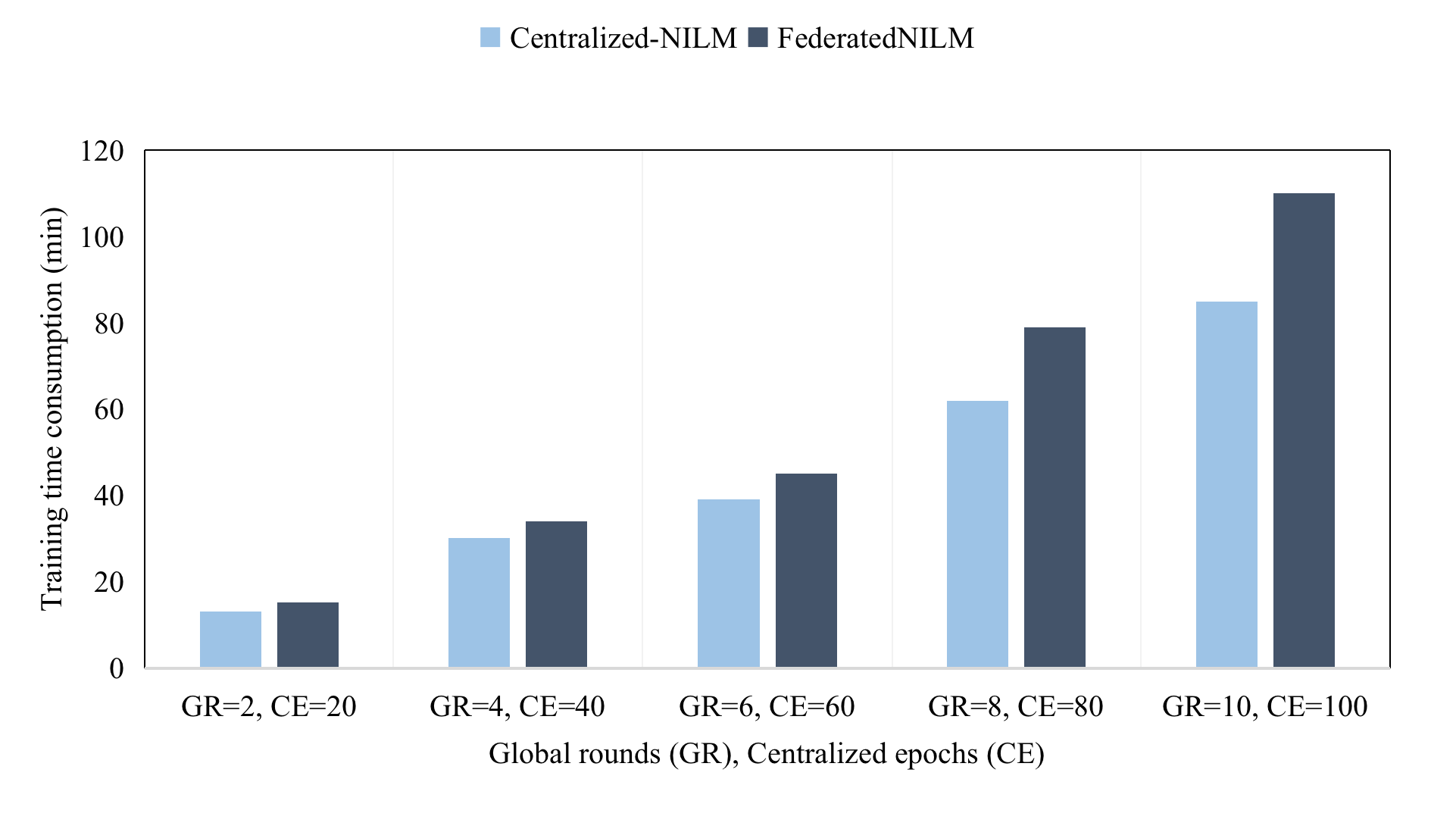}
	\caption{Training time for the proposed FederatedNILM and the centralized NILM model}
	\label{F3}
\end{figure}

The epochs for the Centralized-NILM are set to 20, 40, 60, 80, and 100 respectively corresponding to the global rounds 2, 4, 6, 8, 10 with 10 local epochs of the FederatedNILM model. Tables \ref{t5} to \ref{t7} listed the testing results of both models using different global rounds in the unseen house case for fridge, dishwasher, washing machine respectively, and the best results of each model are marked in bold. Figure. \ref{F3} shows the training time of the two models with the increasing of epochs/global rounds.
\begin{table}[!htbp]
	\centering
	\setlength{\tabcolsep}{0.5mm}{
		\setlength{\abovecaptionskip}{0pt}%
		\setlength{\belowcaptionskip}{0pt}%
		\caption{Test results for fridge}
		\begin{center}
			\begin{tabular}{cllllcllll}
				\toprule
				\hline
				\multirow{2}{*}{\begin{tabular}[c]{@{}l@{}}Global \\ rounds\end{tabular}} &
				\multicolumn{4}{l}{FederatedNILM} &
				\multirow{2}{*}{\begin{tabular}[c]{@{}l@{}}Baseline \\ epochs\end{tabular}} &
				\multicolumn{4}{l}{Centralized-NILM} \\ \cline{2-5} \cline{7-10} 
				& Acc.          & Pre.          & Rec.          & $F_1$         &     & Acc. & Pre. & Rec. & $F_1$ \\ \midrule \hline
				2  & 0.80          & 0.80          & 0.77          & 0.70          & 20  & 0.80 & 0.76 & 0.73 & 0.75  \\ \hline
				4  & 0.82 & 0.69 & 0.76 & 0.74          & 40  & 0.81 & 0.78 & 0.73 & 0.75  \\ \hline
				6  & 0.83          & 0.72          & \textbf{0.86}          & 0.78          & 60  & 0.83 & 0.79 & 0.74 & 0.75  \\ \hline
				8  & 0.84 & 0.70 & 0.82          & 0.73 & 80  & 0.83   & 0.80   & 0.73   & 0.79    \\ \hline
				10 & \textbf{0.85}          & \textbf{0.83}          & 0.79 & \textbf{0.78}          & 100 & \textbf{0.85} & \textbf{0.82} & \textbf{0.77} & \textbf{0.79}  \\ \hline

			\end{tabular}
			\label{t5}
	\end{center}}
\end{table}

\begin{table}[!t]
	\centering
	\setlength{\tabcolsep}{0.5mm}{
		\setlength{\abovecaptionskip}{0pt}%
		\setlength{\belowcaptionskip}{0pt}%
		\caption{Test results for dishwasher}
		\begin{center}
			\begin{tabular}{cllllcllll}
				\toprule
				\hline
				\multirow{2}{*}{\begin{tabular}[c]{@{}l@{}}Global \\ rounds\end{tabular}} &
				\multicolumn{4}{l}{FederatedNILM} &
				\multirow{2}{*}{\begin{tabular}[c]{@{}l@{}}Baseline \\ epochs\end{tabular}} &
				\multicolumn{4}{l}{Centralized-NILM} \\ \cline{2-5} \cline{7-10} 
				& Acc.          & Pre.          & Rec.          & $F_1$         &     & Acc. & Pre. & Rec. & $F_1$ \\ \midrule \hline
				2  & 0.79          & 0.55          & 0.54          & 0.45          & 20  & 0.97 & 0.70 & 0.74 & 0.74  \\ \hline
				4  & 0.81          & 0.58          & 0.57          & 0.68          & 40  & 0.97 & 0.74 & 0.81 & 0.73  \\ \hline
				6  & 0.98          &  0.73          & 0.69          & 0.69          & 60  & 0.98 & 0.74 & 0.84 & 0.77  \\ \hline
				8  & 0.89          & 0.75 & 0.77          & 0.69 & 80  & 0.98   & 0.76   & \textbf{0.91}   & 0.77    \\ \hline
				10 & \textbf{0.98}          & \textbf{0.77}          & \textbf{0.82} & \textbf{0.78}          & 100 & \textbf{0.98} & \textbf{0.76} & 0.90 & \textbf{0.79}  \\ \hline
			\end{tabular}
			\label{t6}
	\end{center}}
\end{table}

\begin{table}[!t]
	\centering
	\setlength{\tabcolsep}{0.5mm}{
		\setlength{\abovecaptionskip}{0pt}%
		\setlength{\belowcaptionskip}{0pt}%
		\caption{Test results for washing machine}
		\begin{center}
			\begin{tabular}{cllllcllll}
				\toprule
				\hline
				\multirow{2}{*}{\begin{tabular}[c]{@{}l@{}}Global \\ rounds\end{tabular}} &
				\multicolumn{4}{l}{FederatedNILM} &
				\multirow{2}{*}{\begin{tabular}[c]{@{}l@{}}Baseline \\ epochs\end{tabular}} &
				\multicolumn{4}{l}{Centralized-NILM} \\ \cline{2-5} \cline{7-10} 
				& Acc.          & Pre.          & Rec.          & $F_1$         &     & Acc. & Pre. & Rec. & $F_1$ \\ \midrule \hline
				2  & 0.96          & 0.52          & 0.53          & 0.41          & 20  & 0.97 & 0.53 & 0.49 & 0.44  \\ \hline
				4  & 0.95          & 0.53          & 0.57          & 0.45          & 40  & 0.98 & 0.50 & 0.52 & 0.47  \\ \hline
				6  & 0.95          & 0.54          & 0.58          & 0.50          & 60  & 0.98 & 0.55 & 0.57 & 0.49  \\ \hline
				8  & 0.94          & 0.55   & 0.58          & \textbf{0.54} & 80  & 0.98   & 0.59   & 0.55   & 0.53    \\ \hline
				10 & \textbf{0.96} & \textbf{0.56}          & \textbf{0.59} & 0.53          & 100 & \textbf{0.98} & \textbf{0.61} & \textbf{0.60} & \textbf{0.54}  \\ \hline
				
			\end{tabular}
			\label{t7}
	\end{center}}
\end{table}

The overall performance of the FederatedNILM model and the Centralized-NILM generally improves with the increase of the epochs.

With the increase of the global rounds, the time consumption cost also increases. The training time consumption of the FederatedNILM model is slightly higher but still within the acceptable range compared with the Centralized-NILM. Besides, given similar classification scores, the proposed FederatedNILM model has similar training time efficiency compared with the Centralized-NILM. For instance, the proposed model obtained an accuracy, precision, recall, $F_1$ of 0.98, 0.77, 0.82, 0.78 respectively for the dishwasher when global rounds reach 10 whereas similar scores (0.98, 0.76, 0.90, 0.79 for accuracy, precision, recall, $F_1$, respectively) were achieved in the Centralized-NILM with 100 epochs. Therefore, we can also reasonably infer that the training cost of FederatedNILM is within an acceptable range compared with the Centralized-NILM under similar accuracy requirements. The above analysis can further confirm the proposed model could achieve a satisfactory trade-off between the computational costs and the model performance.

\subsubsection{Performance comparison with state-of-the-arts }\label{compare}

In the previous subsections, we compare the proposed FederatedNILM with Centralized-NILM to examine its feasibility from the perspective of communication cost and model accuracy (i.e. distributed vs centralized). In the following, we compare our adopted deep learning architecture (used in both  FederatedNILM and Centralized-NILM) with state-of-the-arts considered in \cite{kelly} on model generalization performance for NILM task. 

By considering the experiment environment setup in this paper and in \cite{kelly} and to allow for a direct comparison, the appliances and corresponding house numbers listed in Table \ref{tunseenkelly} are considered. 

\begin{table}[!htbp]
	\centering
	\setlength{\tabcolsep}{0.9mm}{
		\setlength{\abovecaptionskip}{0pt}%
		\setlength{\belowcaptionskip}{10pt}%
		\caption{Training and testing houses for the unseen house case model}
		\begin{center}
			\begin{tabular}{lll}
				\toprule
				\hline
				& Train & Testing \\\midrule 
				\hline
				Dishwasher      & [1, 2]  & 5          \\ \hline
				Washing machine      & [1, 5]  & 2          \\ \hline
			\end{tabular}
			\label{tunseenkelly}
	\end{center}}
\end{table}

Table \ref{t8} gives the comparative results of FederatedNILM and Centralized-NILM with state-of-the-arts for dishwasher and washing machine respectively. The models considered for the comparative study are combinatorial optimization(CO), factorial hidden Markov models (Neural-NILM FHMM), and deep neural network (Autoencoder, Rectangles, Neural-NILM LSTM, centralized-NILM, and  FederatedNILM).

\renewcommand\arraystretch{1}
\begin{table}[!htbp]
	\centering
		\caption{Comparison results of Federated-NILM with state-of-the-arts}
		\begin{center}
			\resizebox{1\textwidth}{!}{
 
            \begin{tabular}{lllllllll}
				\toprule
				\hline
				\multirow{2}{*}{\begin{tabular}[c]{@{}l@{}}State-of-the-art \\ models\end{tabular}}                                    & \multicolumn{4}{c}{Dishwasher}                                & \multicolumn{4}{c}{Washing machine}                           \\
				\cline{2-9}

				& Acc.          & Pre.          & Rec.          & $F_1$         & Acc.          & Pre.          & Rec.          & $F_1$         \\\midrule \hline

				CO \cite{kelly}                                                                                          & 0.64          & 0.06          & 0.67          & 0.11          & 0.88          & 0.06          & 0.48          & 0.10          \\ \hline
			\begin{tabular}[c]{@{}l@{}}Neural-NILM \\ FHMM \cite{kelly}\end{tabular}                                   & 0.33          & 0.03          & 0.49          & 0.05          & 0.79          & 0.04          & 0.64          & 0.08          \\ \hline
				Autoencoder \cite{kelly}                                                                                  & 0.92          & 0.29          & \textbf{0.99} & 0.44          & 0.82          & 0.07          & \textbf{1.00}          & 0.13          \\ \hline
				Rectangles \cite{kelly}                                                                                    & \textbf{0.99}          & \textbf{0.89}          & 0.64          & 0.74          & 0.98          & 0.29          & 0.24          & 0.27          \\ \hline
			
			\begin{tabular}[c]{@{}l@{}}Neural-NILM \\ LSTM\cite{kelly}\end{tabular}                                   & 0.30          & 0.04          & 0.87          & 0.08          & 0.23          & 0.01          & 0.73          & 0.03          \\ \hline
				
				\begin{tabular}[c]{@{}l@{}} Centralized \\-NILM\end{tabular}                                  & \textbf{0.99} & 0.79 & \textbf{0.99}          & \textbf{0.83}        &  \textbf{1.00}          & \textbf{0.85}          & 0.96          & 0.90          \\ \hline

				\begin{tabular}[c]{@{}l@{}} FederatedNILM\end{tabular}                                 & 0.98      & 0.84          & 0.69          & 0.78          & \textbf{1.00}          & 0.77          & 0.67          & \textbf{0.92}          \\ \hline

			\end{tabular}
			}
			\label{t8}
	\end{center}
\end{table}

In general, the deep neural network-based models for NILM have relatively better performance than other models. It is also worth pointing out that both the Centralized-NILM and FederatedNILM achieved better testing results on the unseen house 2 (see Tables \ref{tunseenkelly} and \ref{t8}) than the average testing results of all three cases in Table \ref{tunseen} for the washing machine. One possible reason is that house 1 provides the largest dataset, and when this house is considered as an unseen house as in Table \ref{tunseen}, the training size of the model becomes much smaller, which could impact the model performance. Among all the methods, the Centralized-NILM model achieved the highest accuracy score for both appliances, which proves that the deep learning architecture we utilized in this paper could enhance the local training performance and thus improve the overall state inference accuracy in our framework. As already discussed in previous subsections, the above results also demonstrated that our proposed FederatedNILM with the same deep learning architecture as in Centralized-NILM could achieve promising generalization performance. Considering FederatedNILM works in a distributed and privacy-preserving manner, which is fundamentally different from the other considered methods (i.e. centralized based methods), our proposed framework could achieve a good trade-off between data privacy protection and data utility. 

\section{Conclusion}
In this paper, we proposed a distributed and privacy-preserving framework based on federated learning for NILM (FederatedNILM) to classify the typical states of appliances on the household level. The proposed FederatedNILM combines federated learning with a novel deep neural network model, which could benefit from labelled data of multiple distributed user data sources for NILM. More importantly, compared with the standard centralized model, the framework only requires model parameters uploading instead of data transmitting, which has good capability for data privacy protection. Comparative experiments on a real-world dataset demonstrate the feasibility and good performance of the proposed distributed and privacy-preserving framework as well as the superior model performance of the adopted deep learning architecture for NILM.

Although our proposed FederatedNILM achieved satisfactory results for most of the cases, its average disaggregation performance on the washing machine for the unseen house case still needs to be improved. The same situation has been found in \cite{kelly}. Therefore, in our future work, we will explore to further improve the NILM generalization model performance for particular appliances with more complicated characteristics. In addition, how to further optimize our FederatedNILM framework to enhance its privacy-preserving level with minimal reduction of data utility is another future research direction that is worth exploring.

\section*{Acknowledgement}
This work was supported by the PhD Studentship at University of Essex under the project ``A distributed and real-time learning framework for smart meter big data" (2019-2022).

\bibliography{FDL}

\end{document}